\documentclass[10pt,twocolumn,letterpaper]{article}

\usepackage{cvpr}              

\usepackage[dvipsnames]{xcolor}
\newcommand{\rf}[1]{{\textbf{\color{red}#1}}}
\newcommand{\bd}[1]{{\color{blue}#1}}

\definecolor{cvprblue}{rgb}{0.21,0.49,0.74}
\usepackage[pagebackref,breaklinks,colorlinks,citecolor=cvprblue]{hyperref}
\usepackage{multirow}
\usepackage{pifont}
\def\confName{CVPR}
\def\confYear{2024}

\title{Cascaded Temporal Updating Network for Efficient Video Super-Resolution}

\author{Hao Li \quad Jiangxin Dong \quad Jinshan Pan\\
School of Computer Science and Engineering, Nanjing University of Science and Technology\\
}

\begin{document}
\maketitle
\begin{abstract}
Existing video super-resolution (VSR) methods generally adopt a recurrent propagation network to extract spatio-temporal information from the entire video sequences, exhibiting impressive performance.
However, the key components in recurrent-based VSR networks significantly impact model efficiency, e.g., the alignment module occupies a substantial portion of model parameters, while the bidirectional propagation mechanism significantly amplifies the inference time.
Consequently, developing a compact and efficient VSR method that can be deployed on resource-constrained devices, e.g., smartphones, remains challenging.
To this end, we propose a cascaded temporal updating network (CTUN) for efficient VSR. 
We first develop an implicit cascaded alignment module to explore spatio-temporal correspondences from adjacent frames.
Moreover, we propose a unidirectional propagation updating network to efficiently explore long-range temporal information, which is crucial for high-quality video reconstruction.
Specifically, we develop a simple yet effective hidden updater that can leverage future information to update hidden features during forward propagation, significantly reducing inference time while maintaining performance.
Finally, we formulate all of these components into an end-to-end trainable VSR network.
Extensive experimental results show that our CTUN achieves a favorable trade-off between efficiency and performance compared to existing methods.
Notably, compared with BasicVSR, our method obtains better results while employing only about 30\% of the parameters and running time.
The source code and pre-trained models will be available at \href{https://github.com/House-Leo/CTUN}{https://github.com/House-Leo/CTUN}.
\end{abstract}

\section{Introduction}
Video Super-Resolution (VSR) aims to reconstruct high-resolution (HR) video sequences by extrapolating missing details from their corresponding low-resolution (LR) inputs.
With the rapid development of streaming and high-definition devices, such as 4K live streaming, TikTok short videos, iPhone 14 Pro ($2532\times1170$), and iPad Pro ($2732\times2048$), the demand for HR videos is steadily rising.
Thus, developing an efficient and effective VSR method that can be deployed on these resource-constrained platforms or devices is of paramount importance.

\begin{figure}[t]
	\centering
	\includegraphics[width=0.46\textwidth]{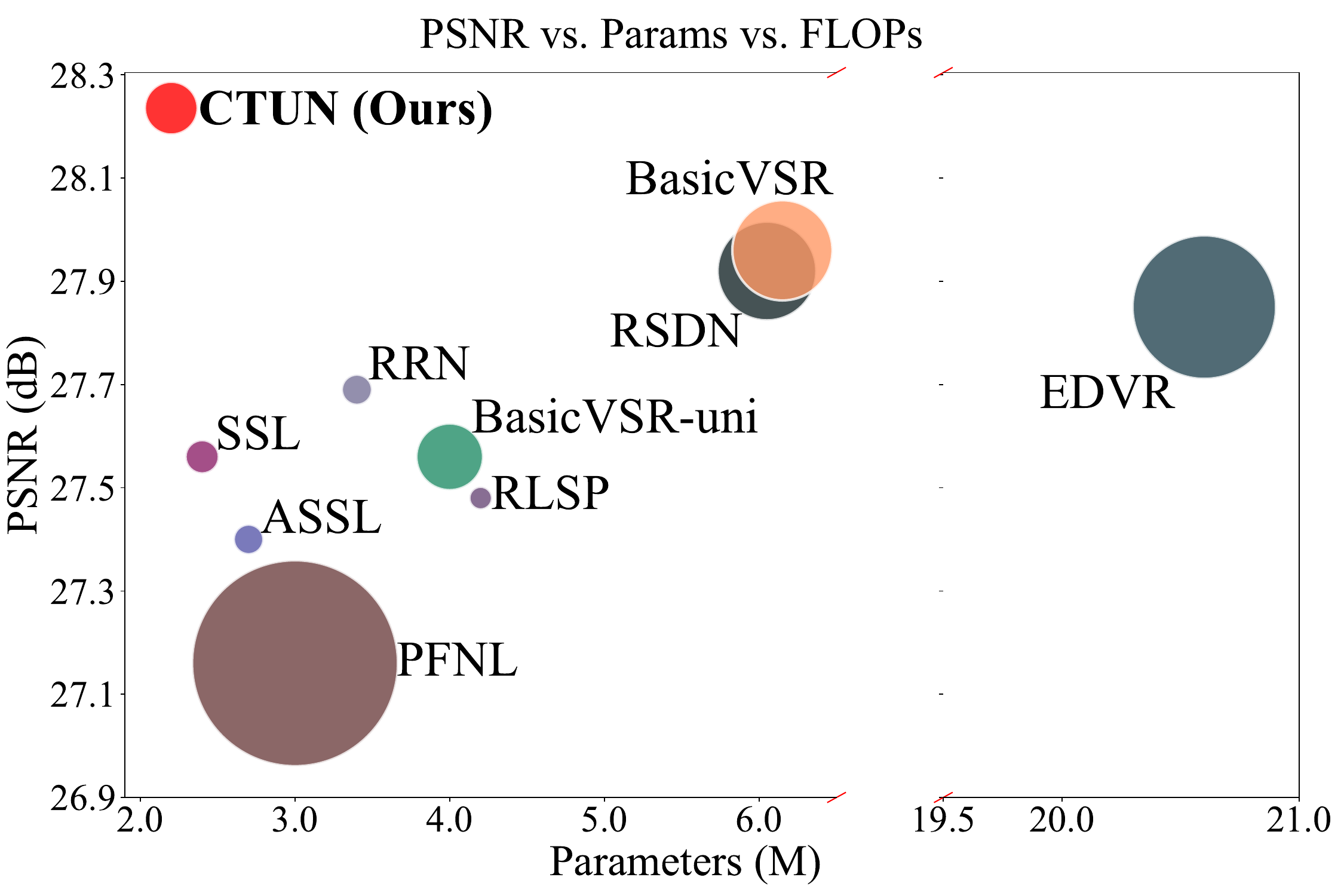}
    \includegraphics[width=0.46\textwidth]{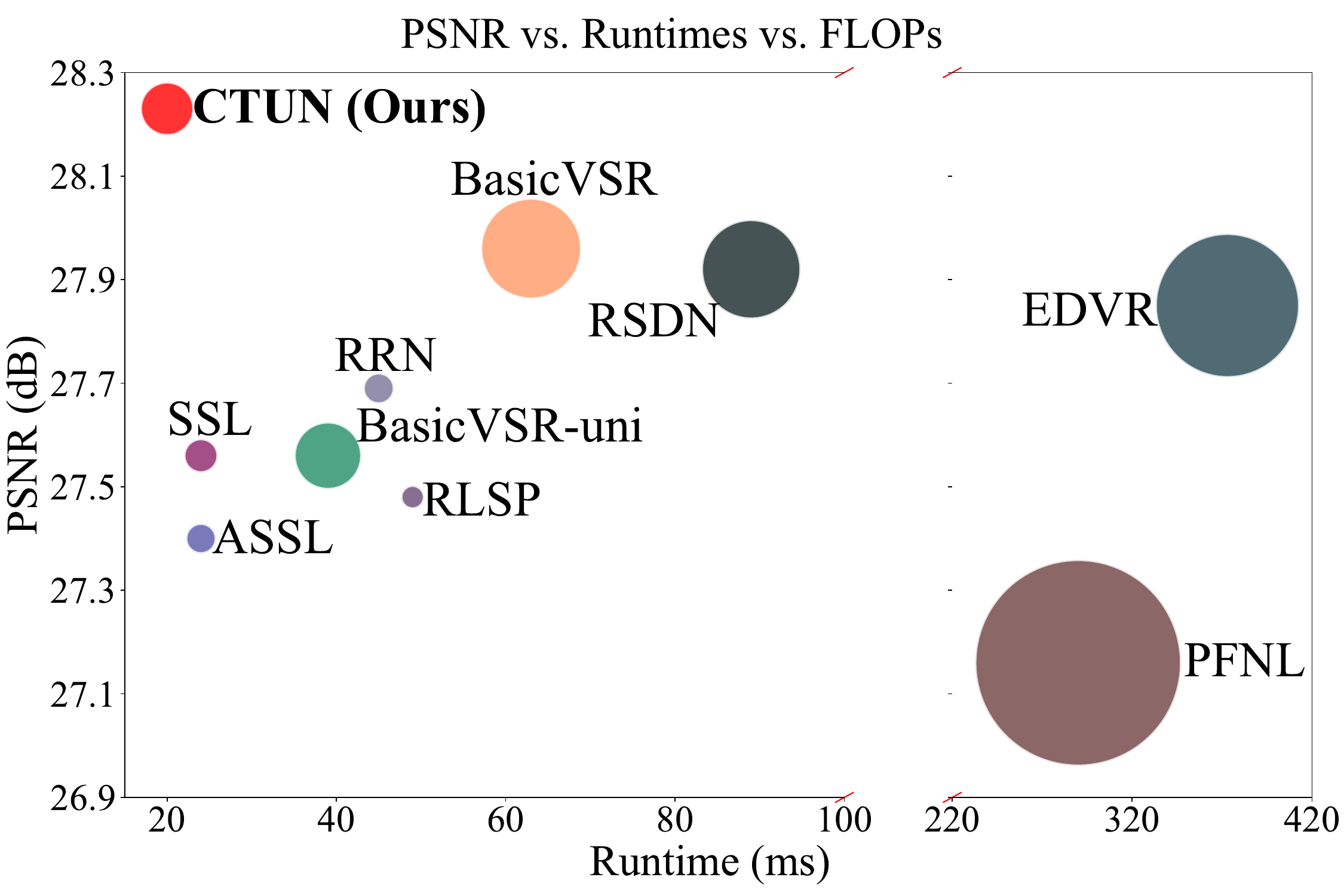}
	\vspace{-1mm}
	\caption{Quantitative comparisons of our proposed model and other state-of-the-art methods on the Vid4~\cite{liu2014bayesian} dataset in terms of PSNR, model parameters, running time, and FLOPs. Circle sizes indicate the number of FLOPs. The proposed CTUN achieves a better trade-off between efficiency and performance.}  
	\label{fig:comp}
	\vspace{-4mm}
\end{figure}

Different from single image super-resolution (SISR) that only considers spatial information of a single image, VSR is more challenging as it needs to take full advantage of spatio-temporal information between multi-frames. 
In recent years, various VSR methods utilize a sliding-window mechanism to handle multi-frame inputs.
For instance, EDVR~\cite{wang2019edvr} and TDAN~\cite{tian2020tdan} adopt deformable convolution for feature alignment and integrate the features from different frames.
However, these methods based on local information propagation inevitably increases runtime and computational costs, and neglects long-range temporal information, limiting their performance.

To solve this problem, some recurrent-based methods~\cite{chan2021basicvsr,chan2022basicvsrplusplus,liang2022rvrt,liang2022vrt,shi2022rethinking} introduce a propagation manner to explore long-range temporal information by a recurrent propagation network, which consists of an alignment module and a propagation module.
However, the efficiency of recurrent-based VSR methods has been primarily restricted by two key factors: model complexity of the alignment module (\eg, optical flow~\cite{chan2021basicvsr} and deformable convolution~\cite{chan2022basicvsrplusplus}), and the iteration of propagation.
For instance, the alignment modules in BasicVSR~\cite{chan2021basicvsr} and BasicVSR++~\cite{chan2022basicvsrplusplus} account for 23\% (1.44M/6.29M) and 47\% (3.47M/7.32M) of the total parameters, respectively.
Furthermore, each iteration of the recurrent propagation requires traversing the entire video sequences, such as BasicVSR++ (with two bidirectional propagation) is approximately 20\% slower in terms of inference time compared to BasicVSR (with one bidirectional propagation).
Although significant efforts have been made to improve the efficiency of VSR models~\cite{fuoli_rlsp_2019,Xia2022ssl,Xiao_Fu_Huang_Cheng_Xiong_2021},
most of them either need the alignment module with heavy model complexity or require complex training settings (\eg, structured pruning~\cite{Xia2022ssl} and knowledge distillation~\cite{Xiao_Fu_Huang_Cheng_Xiong_2021}), which does not achieve a good trade-off between model efficiency and reconstruction performance.
Therefore, it is of great need to develop a compact VSR model with better trade-off between model efficiency and performance.

In this paper, we develop a cascaded temporal updating network for efficient video super-resolution.
To make model compact, we first propose an implicit cascaded alignment module (ICAM), to align the current frame with its adjacent frames.
Our ICAM enables the cascaded extraction of spatio-temporal information from three time-step features to achieve alignment, instead of using alignment modules with huge model complexity. 
Although the proposed ICAM can extract the spatio-temporal information from the adjacent frames, it has limited capability in exploring long-range temporal information.

To solve this problem, we develop a unidirectional propagation updating network, which consists of a forward propagation module and a hidden updater (HU), to better explore long-range temporal information from video sequences efficiently.
The proposed HU can leverage the information from future frames to update hidden features, thereby introducing bidirectional information into unidirectional propagation for high-quality video reconstruction.
Finally, we formulate the ICAM and the HU into an end-to-end trainable unidirectional propagation network for efficient VSR, named as CTUN.
Experimental results demonstrate that the proposed CTUN achieves comparable performance against state-of-the-art (SOTA) methods on various VSR benchmark datasets.
Figure~\ref{fig:comp} presents the trade-off between model complexity and performance for SOTA VSR methods on the Vid4~\cite{liu2014bayesian} dataset.
In particular, our method outperforms BasicVSR~\cite{chan2021basicvsr} with only about 30\% parameters and running time.

The main contributions are summarized as follows:
\begin{itemize}
    \item  We propose an implicit cascaded alignment module to effectively explore the spatio-temporal correspondences of past, current and future features in a cascaded manner, making model parameter-efficient and easy for training.
    \item We develop a hidden updater that leverages future information to update hidden features during forward propagation, significantly reducing the memory consumption and inference time of recurrent-based VSR models while maintaining performance.
    \item We quantitatively and qualitatively evaluate the proposed method on VSR benchmark datasets. The results demonstrate that our method achieves a favorable trade-off between performance and model complexity.
\end{itemize}

\begin{figure*}[t]
	\centering
	\includegraphics[width=1.0\textwidth]{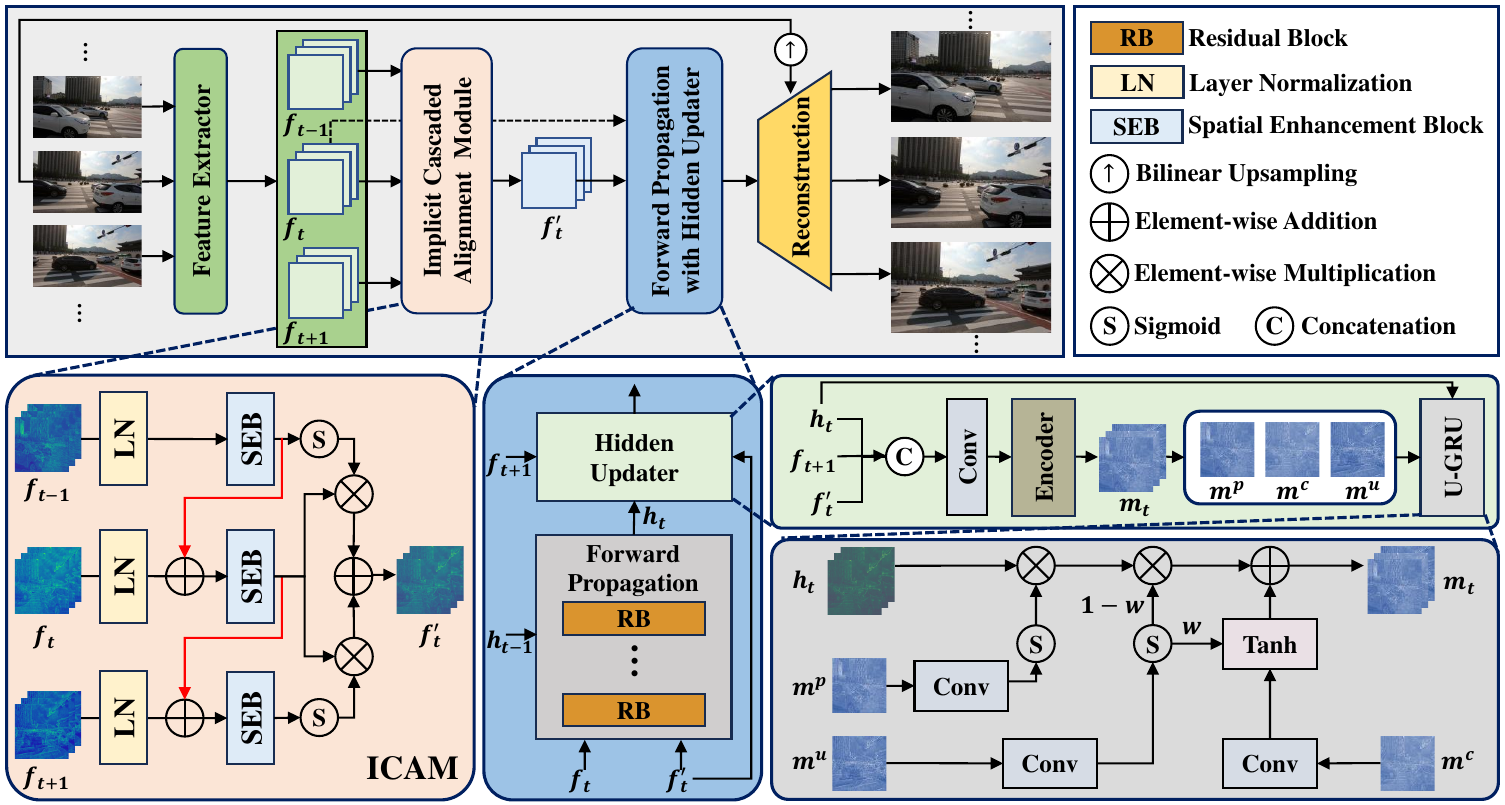} 
	\vspace{-5mm}
	\caption{The overall architecture of our proposed CTUN. The proposed CTUN contains a feature extractor, an implicit cascaded alignment module (ICAM), a forward propagation module with a hidden updater (HU), and a reconstruction module. The ICAM can explore spatio-temopral correspondences between the current features and adjacent features for alignment. Moreover, the HU utilizes a feature encoder to extract spatial information, and employs an updating GRU (U-GRU) to achieve temporal interaction. Here $h_t$ denotes the $t$-th timestep hidden features. And $\{m^p,m^c,m^u\}$ are the past, current, and future components of the features $m_t$, respectively.}
	\vspace{-5mm}
	\label{fig:network}
\end{figure*}

\section{Related Work}
\noindent \textbf{Recurrent-based video super-resolution.} Recurrent-based video super-resolution methods~\cite{liang2022rvrt,liang2022vrt,shi2022rethinking,isobe2022look} exploit long-range temporal information using recurrent propagation networks.
FRVSR~\cite{sajjadi2018frame} recurrently propagates the restored HR video frames to obtain the final HR outputs.
RSDN~\cite{isobe2020video} proposes a recurrent structure-detail block and a hidden-state adaptation module for unidirectional propagation, to improve the performance and robustness of the model.
However, these unidirectional propagation methods only utilize the past estimated results to current features for reconstruction.
To address this problem, BasicVSR~\cite{chan2021basicvsr} adopts a bidirectional temporal propagation that propagates video sequences forward and backward in time independently, and proposes flow-based feature alignment to achieve a significant improvement.
BasicVSR++~\cite{chan2022basicvsrplusplus} modifies BasicVSR by employing second-order grid propagation and flow-guided deformable alignment, achieving state-of-the-art performance in the VSR task.
Similarly, Yi \etal~\cite{Yi_2021_ovsr} propose a omniscient VSR network to estimate hidden states from the past, current, and future LR frames.
Although these recurrent-based methods achieve promising performance, they rely on motion compensation provided by heavy alignment modules, and they require a huge memory consumption due to the storage of hidden features at each propagation stage.
Different from previous methods, we focus on developing an efficient recurrent-based VSR method with lower memory consumption for high-quality video reconstruction.

\noindent \textbf{Efficient video super-resolution.} In contrast to SISR, VSR requires the extraction of temporal interactions between inter-frames, making commonly used methods utilized for efficient SISR, such as structured pruning~\cite{zhang2021aligned,zhang2022learning}, structural re-parameterization~\cite{Zhang2021ecbsr}, knowledge distillation~\cite{he2020_fakd}, unsuitable for direct application to VSR models.
RLSP~\cite{fuoli_rlsp_2019} introduces high-dimensional latent states to propagate temporal information implicitly, avoiding processing LR frames multiple times.
Xiao \etal~\cite{Xiao_Fu_Huang_Cheng_Xiong_2021} propose a space-time knowledge distillation scheme to exploit spatial and temporal knowledge in VSR models, which improves performance, but does not effectively reduce inference time.
Xia \etal~\cite{Xia2022ssl} develop a structured pruning scheme called structured sparsity learning (SSL) to prune the unimportant filters of recurrent-based VSR networks.
Although it is able to make model compact, it still needs the heavy alignment module for motion estimation, and inaccurate pruning may cause performance degradation.
Some public challenges~\cite{Ignatov_2022_challenge,Ignatov_Romero2021challenge} aim to develop end-to-end real-time VSR methods for mobile NPUs deployment.
However, to improve model efficiency, the winning solution in~\cite{Ignatov_2022_challenge} still employs SISR methods for VSR data, disregarding temporal information and limiting performance.
Different from these methods, we aim to develop an efficient VSR method with a compact alignment module that achieves a favorable trade-off between model efficiency and performance.

\vspace{-4mm}
\section{Proposed Method}
\vspace{-1mm}
Our goal is to present an efficient and effective method to explore spatio-temporal information for high-quality video reconstruction.
To this end, we first develop an implicit cascaded alignment module (ICAM) to efficiently capture spatial and temporal correspondences between the current features and adjacent features for alignment.
To extract long-range temporal information from video sequences, we further develop a unidirectional propagation network with a hidden updater (HU), which can not only improve the performance but also significantly reduce the inference time.
We formulate these components into an end-to-end trainable propagation network for efficient VSR, called cascaded temporal updating network (CTUN).
In the following, we present the details of our methods.

\subsection{Overview}
As shown in Figure~\ref{fig:network}, our proposed CTUN consists of a feature extractor, an implicit cascaded alignment module (ICAM), a forward propagation module with a hidden updater (HU), and a reconstruction module.
The goal is to super-resolve a LR video clips $X=\{x_0,x_1,\cdots,x_{N-1}\}$ with $N$ frames to a HR video clips $Y=\{y_0,y_1,\cdots,y_{N-1}\}$ by a given scaling factor $s$, where $y_t \in \mathbb{R}^{C \times sH \times sW}$ denotes the reconstructed HR frame of the corresponding LR frame $x_t \in \mathbb{R}^{C \times H \times W}$ at timestep $t$, $H$ and $W$ are the spatial resolution and $C$ is the number of channels.

We first embed all video frames into the same feature space by applying the feature extractor to each input frame, and the embedded features are represented as $F=\{f_0,f_1,\cdots,f_{N-1}\}$.
In contrast to aligning features by explicitly estimating optical flow, we employ an implicit method to progressively align the features at the current timestep.
Specifically, at timestep $t$, we use $\{f_{t-1},f_t,f_{t+1}\}$ as the input for ICAM because adjacent frames usually contain the most relevant spatio-temporal information to the current frame.
To align the current feature $f_t$, ICAM progressively propagates the information from $f_{t-1}$ to $f_{t+1}$, and utilizes a gating mechanism to integrate the most useful information into $f_t$ and obtain the aligned feature $f'_t$.

After obtaining the aligned feature $f'_t$, we concatenate it with $f_t$ and input them into the forward propagation module, resulting in the hidden feature $h_t$.
As $h_t$ in the forward propagation lacks information from future frames, we leverage $f'_t$ containing such information, and employ a HU to update $h_t$ to mitigate the performance drop due to the unidirectional propagation.
Finally, we restore the HR video clip by concatenating $f_t$, $f'_t$ and $h_t$ using the reconstruction module and the upsampling operation.

\vspace{-1mm}
\subsection{Implicit cascaded alignment module}
One of the major procedures in existing SOTA VSR methods is to align the adjacent frames, enabling the fusion of corresponding pixels and enhancing the reconstruction quality of the current frame.
The most commonly used alignment methods include optical flow~\cite{tao2017detail,chan2021basicvsr} and deformable convolution~\cite{wang2019edvr,tian2020tdan}, which can estimate the offset between two frames and align them using the warping operation.
However, as the explicit interpolation (\textit{e.g.,} bilinear) in the warping operation needs to re-sample the images, most of these offset estimation methods are designed for high-quality images, it is a challenging task to estimate accuracy results in low-quality frames.
As a consequence, the inaccurate alignment has a negative impact on the performance of VSR.

To mitigate this issue, we propose an implicit cascaded alignment module (ICAM) to effectively and efficiently explore the temporal correspondences between the current frame and its adjacent frames.
The proposed ICAM consists of three LayerNorm~\cite{ba2016layer} layers and three spatial enhancement blocks (SEB) (see Figure~\ref{fig:network}), each corresponding to three different timestep inputs.
Specifically, at timestep $t$, we use consecutive features $\{f_{t-1},f_t,f_{t+1}\}$ as the ICAM inputs, we first normalize three inputs respectively, and then extract spatio-temporal information in a cascaded manner, the procedure is formulated as:
\vspace{-1mm}
\begin{equation}
\vspace{-1mm}
    \begin{split}
         &\{f_{t-1},f_t,f_{t+1}\} = \text{LN}(\{f_{t-1},f_t,f_{t+1}\}), \\
         &f_{i} = \mathcal{F}(f_i+f_{i-1}), ~~~~i \in \{t-1,t,t+1\}, 
    \end{split}
    \label{eq:ICAM}
\end{equation}
where $\text{LN}(\cdot)$ and $\mathcal{F}(\cdot)$ denote the layer normalization operation and the function of SEB.
When $i=t-1$, $f_{i-1}=0$.
The details of SEB are included in the supplemental material.

As we desire to select discriminative features by incorporating both past and future temporal information simultaneously, we utilize a gating mechanism to fusion these three timesteps features of \eqref{eq:ICAM}.
It can be expressed by:
\vspace{-1mm}
\begin{equation}
\vspace{-1mm}
    f'_t = \sigma(f_{t-1}) \otimes f_t + \sigma(f_{t+1}) \otimes f_t,
    \label{eq:Gate}
\end{equation}
where $\sigma(\cdot)$ and $\otimes$ denote the Sigmoid function and the element-wise multiplication.

Due to the cascaded design of ICAM, the spatio-temporal information can be propagated from timestep $t-1$ to $t+1$, allowing the features of current timestep to simultaneously incorporate information from past and future frames for alignment.
And the gating mechanism maintains a good balance between the past and future information for fusing them to current features.

As the proposed ICAM only considers the spatio-temporal information from adjacent frames, we further develop a unidirectional propagation network with a hidden updater to explore long-range temporal information for high-quality video reconstruction.

\subsection{Hidden updater}
The introduction of bidirectional propagation~\cite{chan2021basicvsr,chan2022basicvsrplusplus,liang2022rvrt} achieves a significant improvement in the VSR task.
However, it demands looping through the entire video sequences twice and storing forward and backward hidden features, resulting in huge memory consumption and a substantial increase in inference time.

To reduce inference time and enhance model efficiency, we propose a unidirectional propagation (\ie, forward propagation) network to explore long-range temporal information efficiently.
The forward propagation module can extract spatio-temporal correspondence from the concatenated input of shallow features $f_t$, aligned features $f'_t$, and previous hidden features $h_{t-1}$:
\vspace{-1mm}
\begin{equation}
\vspace{-1mm}
    h_t = \mathcal{H}_p(\text{Concat}(f_t, f'_t, h_{t-1})),
    \label{eq:forward}
\end{equation}
where $\mathcal{H}_p(\cdot)$ and $\text{Concat}$ denote the forward propagation module and channel-wise concatenation operation, respectively.
Although unidirectional propagation has faster inference time compared to bidirectional propagation, they suffer from poorer performance due to the neglect of future information.
This motivates us to introduce the future information in unidirectional propagation.

To this end, we develop a simple yet effective hidden updater to bring future information to hidden features during forward propagation.
Our hidden updater (HU) consists of a feature encoder and an updating GRU (U-GRU) (see Figure~\ref{fig:network}).
Specifically, to better explore spatial information, we first concatenate $\{f'_t, f_{t+1}\}$ contained future information with hidden features $h_t$, and use it as input for encoder, which are formulated as:
\vspace{-1mm}
\begin{equation}
\vspace{-1mm}
    \begin{split}
        m_t &= \text{Concat}(f_{t+1},h_t,f'_t), m_t \in \mathbb{R}^{3C \times H \times W}, \\
        m_t &= \text{Conv}_{1 \times 1}(m_t), m_t \in \mathbb{R}^{C \times H \times W}, \\
        m_t &= \mathcal{H}_{e}(m_t), m_t \in \mathbb{R}^{C \times H \times W}, \\
    \end{split}
    \label{eq:CBAM}
\end{equation}
where $\text{Conv}_{1 \times 1}$ denotes the $1 \times 1$ convolution, and $\mathcal{H}_{e}(\cdot)$ is the feature encoder module (the details are included in the supplemental material). 

After obtaining the refined representation $m_t$ by \eqref{eq:CBAM}, to introduce more temporal interactions to the hidden feature $h_t$, we utilize the U-GRU to update $h_t$ and further explore temporal information.
As $m_t$ has the cascaded temporal information from $f'_t$, we take $m_t$ and $h_t$ as the U-GRU input.
Specifically, we utilize a $1 \times 1$ convolution to expand $m_t \in \mathbb{R}^{C \times H \times W}$ to $m_t \in \mathbb{R}^{3C \times H \times W}$, and split it into three components $\{m^p,m^c,m^u\}$ from different timesteps, corresponding to past, current, and future features.
The entire procedure is formulated as:
\vspace{-1mm}
\begin{equation}
\vspace{-1mm}
    \begin{split}
        &m_t = \text{Conv}_{1 \times 1}(m_t), \\
        &\{m^p,m^c,m^u\} = \text{Split}(m_t), \\
        &z = \sigma(\text{Conv}_{3 \times 3}(m^p)), \\
        &w = \sigma(\text{Conv}_{3 \times 3}(m^u)), \\
        &q = \text{tanh}(\text{Conv}_{3 \times 3}(m^c)),\\ 
        &m_t = z \otimes h_t \otimes (1-w) + q \otimes w,
    \end{split}
    \label{eq:update}
\end{equation}
where $\text{Conv}_{3 \times 3}(\cdot)$ is the $3 \times 3$ convolution, $q$ represents the values of the current features, and $z$ and $w$ are the scale factor of information from the past and future, respectively.
With \eqref{eq:update}, the original hidden features $h_t$ can be updated using learnable scale factors, to introduce future information in the temporal domain during forward propagation.
That is, when $t>0$, we use the updated hidden feature $m_t$ as input to the forward propagation, and \eqref{eq:forward} changes as follows:
\vspace{-1mm}
\begin{equation}
    h_t = \mathcal{H}_p(\text{Concat}(f_t, f'_t, m_{t-1})).
    \label{eq:newforward}
\end{equation}

After propagation, we can restore the HR video frame $y_t$ by aggregating the shallow, aligned and propagation features using a reconstruction module which contains three residual blocks~\cite{he2016deep} and two pixel-shuffle layers~\cite{shi2016real}, which is formulated as:
\vspace{-1mm}
\begin{equation}
\vspace{-1mm}
    y_t = \mathcal{R}(\text{Concat}(f_t, f'_t, h_t)) + \mathcal{U}(x_t),
\end{equation}
where $\mathcal{R}(\cdot)$ denotes the reconstruction module, and $\mathcal{U}(\cdot)$ denotes the bilinear upsampling operation.

\begin{table*}[!ht]
    \caption{Quantitative comparisons (PSNR/SSIM) with state-of-the-art methods for VSR ($\times 4$). \rf{Red} and \bd{Blue} indicate the best and the second-best performance, respectively. Where `uni' and `bi' represents the unidirectional and bidirectional propagation. The FLOPs and runtime are the average results computed based on 100 LR video frames with a size of $180{\times}320$. All results are calculated on Y-channel except REDS4~\cite{nah2019ntire}.}
    \label{tab:quan}
    \vspace{-2mm}
    \resizebox{1.0\textwidth}{!}{
            \tabcolsep=0.1cm
            \begin{tabular}{l||ccc||c|c|c||c|c|c}
                \toprule
                \multirow{2}{*}{Methods}     & \multirow{2}{*}{Params. (M)}    & \multirow{2}{*}{FLOPs (G)}         & \multirow{2}{*}{Runtime (ms)}      & \multicolumn{3}{c||}{BI degradation}                                                                    & \multicolumn{3}{c}{BD degradation}                 \\ \cline{5-10}
                           & & &   & REDS4~\cite{nah2019ntire}          & Vimeo-T~\cite{xue2019video}      & Vid4~\cite{liu2014bayesian} & UDM10~\cite{yi2019progressive} & Vimeo-T~\cite{xue2019video}   & Vid4~\cite{liu2014bayesian}          \\ \hline \hline
                Bicubic           & -          & -   &-         & 26.14/0.7292     & 31.32/0.8684                       & 23.78/0.6347                & 28.47/0.8253                   & 31.30/0.8687                    & 21.80/0.5246                           \\
                TOFlow~\cite{xue2019video}     & 1.4          & 275   & 1610                        & 27.98/0.7990     & 33.08/0.9054                       & 25.89/0.7651                & 36.26/0.9438                   & 34.62/0.9212                    & -                                    \\
                DUF~\cite{jo2018deep}       & 5.8          & 1646   & 974                           & 28.63/0.8251    & -                                  & -                           & 38.48/0.9605                   & 36.87/0.9447                    & 27.38/0.8329                       \\
                RBPN~\cite{haris2019recurrent}   & 12.2          & 8516   & 1507                     & 30.09/0.8590      & 37.07/0.9435                       & 27.12/0.8180                & 38.66/0.9596                   & 37.20/0.9458                    & -                                 \\
                EDVR~\cite{wang2019edvr}       & 20.6          & 516   & 378                         & 31.09/0.8800   & \rf{37.61/0.9489}                       & 27.35/0.8264                & \bd{39.89/0.9686}                   & \rf{37.81/0.9523}                    & 27.85/0.8503                        \\
                PFNL~\cite{yi2019progressive}   & 3.0    & 940   & 295        & 29.63/0.8502                  & 36.14/0.9363                       & 26.73/0.8029                & 38.74/0.9627                   & -                               & 27.16/0.8355                         \\
                RLSP~\cite{fuoli_rlsp_2019}    & 4.2    & 82   & 49                          & - & -                      & -                & 38.48/0.9606                   & 36.49/0.9403                               & 27.48/0.8388                         \\
                RSDN~\cite{isobe2020video}    & 6.2    & 355   & 94                          & - & -                      & -                & 39.35/0.9653                   & 37.23/0.9471                               & 27.92/0.8505                         \\
                RRN~\cite{isobe2020revisiting}    & 3.4    & 109   & 45                        & -  & -                      & -                & 38.96/0.9644                   & -                              & 27.69/0.8488                         \\
                ASSL~\cite{Zhang_assl_2021}       & 2.7    & 107   & 24        & 30.74/0.8770                  & 36.75/0.9414                      & 27.01/0.8176                & 39.15/0.9660                   & 36.93/0.9450                              & 27.40/0.8400                         \\
                SSL~\cite{Xia2022ssl}             & 2.4    & 120   & 24      & 31.06/0.8933                    & 36.82/0.9419                      & 27.15/0.8208                & 39.35/0.9665                   & 37.06/0.9458                              & 27.56/0.8431                         \\
                BasicVSR-uni~\cite{chan2021basicvsr} & 4.0    & 240    & 39       & 30.56/0.8698        & 36.95/0.9429                       & 27.01/0.8164                & 39.25/0.9645                   & 37.25/0.9472                    & 27.57/0.8424                          \\
                BasicVSR~\cite{chan2021basicvsr} & 6.3    & 364    & 63      & \rf{31.42}/\bd{0.8909}         & 37.18/0.9450                       & 27.24/0.8251                & 39.96/0.9694                   & 37.53/0.9498                    & 27.96/0.8553                          \\
                    
                \hline \hline
                \textbf{CTUN (Ours)}      & 2.2     & 190   & 21     & 31.19/0.8821        & 37.24/0.9461                       & \bd{27.48/0.8283}                & 39.82/0.9683                   & 37.58/0.9511                    & \bd{28.25/0.8592}      \\
                \textbf{CTUN-bi (Ours)}      & 2.8     & 248   & 38     & \bd{31.40}/\rf{0.8911}        & \bd{37.48/0.9470}                       & \rf{27.52/0.8299}                & \rf{40.01/0.9695}                   & \bd{37.72/0.9520}                    & \rf{28.30/0.8600} \\
                \bottomrule
            \end{tabular}}
        \vspace{-2mm}
\end{table*}

\section{Experimental Results}
\subsection{Datasets and implementation details}
\noindent \textbf{Training datasets and benchmarks.}
Following~\cite{wang2019edvr,tian2020tdan,chan2021basicvsr}, we use the commonly used VSR datasets, including REDS~\cite{nah2019ntire} and Vimeo-90K~\cite{xue2019video} for training.
To evaluate our method, we use REDS4~\cite{nah2019ntire}, Vid4~\cite{liu2014bayesian}, Vimeo-T~\cite{xue2019video} and UDM10~\cite{yi2019progressive} as benchmarks, including two degradations - Bicubic (BI) and Blur Downsampling (BD).
For BI, we use the MATLAB function `imresize' for downsampling, while we utilize a Gaussian filter ($\sigma=1.6$) to blur the HR frames for BD. 
All models are trained and tested for $4 \times$ super-resolution, and we adopt PSNR and SSIM as the evaluation metrics.

\noindent \textbf{Implementation details.} We utilize the Adam~\cite{kingma2014adam} optimizer with $[\beta_1,\beta_2]=[0.9,0.99]$ during training. 
The batch size is set to be 32 and the patch size of input LR frames is $64 \times 64$, and the feature number is 64.
The initial learning rate is set to $2 \times 10^{-4}$, and we use cosine annealing scheme~\cite{loshchilov2016sgdr} to decrease the learning rate.
The number of iterations to train our model is set to be $3\times10^6$.
The number of residual blocks in feature extractor, forward propagation module and reconstruction module are set to be $\{3,5,3\}$, respectively.
We use Charbonnier~\cite{charbonnier1994two} and FFT~\cite{cho2021rethinking} loss to constrain the model training.
All experiments are conducted on a server with 8 NVIDIA RTX 3090 GPUs and PyTorch 1.14.
More experimental settings and the details of the network architecture are included in the supplementary material due to the page limit.
The source code and pre-trained models will be available at \href{https://github.com/House-Leo/CTUN}{CTUN}.

\begin{figure*}[t]
	\footnotesize
	\begin{center}
		\begin{tabular}{c c c c c c c c}
			\multicolumn{3}{c}{\multirow{5}*[55pt]{\hspace{-1.5mm}\includegraphics[width=0.49\linewidth, height=0.268\linewidth]{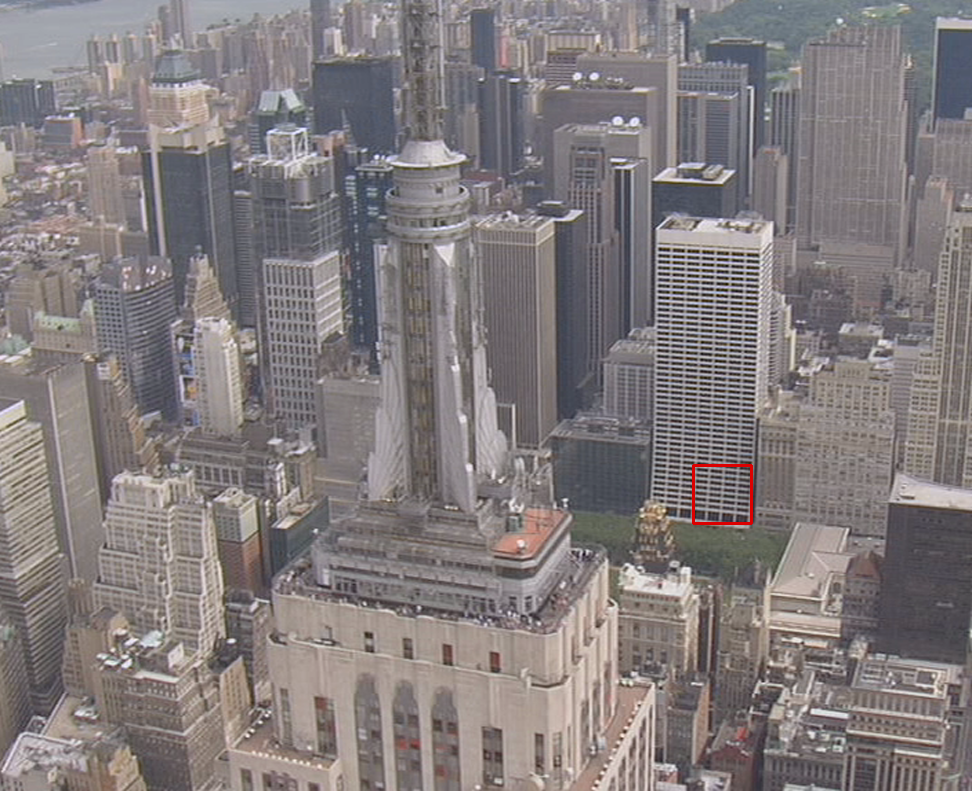}}}
            & \hspace{-4.0mm} \includegraphics[width=0.12\linewidth]{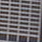} 
            & \hspace{-5.0mm} \includegraphics[width=0.12\linewidth]{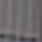}  
            & \hspace{-5.0mm}  \includegraphics[width=0.12\linewidth]{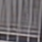}  
            & \hspace{-3.5mm} \includegraphics[width=0.12\linewidth]{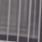}  \\
			\multicolumn{3}{c}{~}                                   &\hspace{-4.0mm} (a)  GT patch               & \hspace{-5.0mm} (b) Bicubic                    & \hspace{-5.0mm} (c) PFNL~\cite{yi2019progressive}  & \hspace{-3.5mm} (d) EDVR~\cite{wang2019edvr} \\		
			\multicolumn{3}{c}{~} 
            & \hspace{-4.0mm} \includegraphics[width=0.12\linewidth]{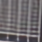} 
            & \hspace{-5.0mm} \includegraphics[width=0.12\linewidth]{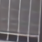} 
            & \hspace{-5.0mm} \includegraphics[width=0.12\linewidth]{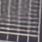} 
            & \hspace{-3.5mm} \includegraphics[width=0.12\linewidth]{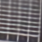}  \\
			\multicolumn{3}{c}{\hspace{-4.0mm} Frame \textit{011}, Clip \textit{City}}&\hspace{-4.0mm} (e) SSL~\cite{Xia2022ssl} & \hspace{-5.0mm} (f) BasicVSR-uni~\cite{chan2021basicvsr}  & \hspace{-5.0mm} (g) BasicVSR~\cite{chan2021basicvsr}          & \hspace{-3.5mm}(h) CTUN (Ours) \\			
		\end{tabular}
	\end{center}
	\vspace{-6mm}
	\caption{Visual comparisons on the Vid4 (BD) dataset. Our method can restore much clearer structure information of the building.}
	\label{fig:visual1}
	\vspace{-5mm}
\end{figure*}

\subsection{Quantitative and qualitative comparisons}
Following~\cite{Xia2022ssl}, we fisrt compare the proposed method with some lightweight VSR methods, including TOFlow~\cite{xue2019video}, DUF~\cite{jo2018deep}, PFNL~\cite{yi2019progressive}, RLSP~\cite{fuoli_rlsp_2019}, RSDN~\cite{isobe2020video}, RRN~\cite{isobe2020revisiting}, BasicVSR-uni and BasicVSR~\cite{chan2021basicvsr}.
In addition, we evaluate the performance of our proposed methods with some efficient VSR methods used structured pruing, including ASSL~\cite{Zhang_assl_2021} and SSL\cite{Xia2022ssl}.
Moreover, we compare the experiment results with some heavy VSR methods, including EDVR~\cite{wang2019edvr} and RBPN~\cite{haris2019recurrent}.

\noindent \textbf{Quantitative comparisons.} Table~\ref{tab:quan} shows the quantitative results.
Notably, our CTUN achieves state-of-the-art performance on Vid4~\cite{liu2014bayesian} and the second-best results on Vimeo-T~\cite{xue2019video}, including two different degradation.
Compared to BasicVSR~\cite{chan2021basicvsr}, our CTUN achieves a PSNR gain of 0.24dB and 0.29dB on Vid4(BI) and Vid4(BD), respectively, while the number of the proposed model parameter is about 34\% of BasicVSR.
Also, we change the original BasicVSR to unidirectional propagation, as BasicVSR-uni, and Table~\ref{tab:quan} represents that our method achieves better performance in all benchmarks.
Furthermore, compared to the recently proposed efficient VSR method, SSL~\cite{Xia2022ssl}, the PSNR gain of our method is at least 0.35dB higher than SSL.
Even compared to the large model, EDVR~\cite{wang2019edvr}, our method achieves comparable results with fewer parameters and lower inference time.
As evidenced by~\cite{Xia2022ssl}, the bidirectional propagation mechanism can handle video sequences with large motion (\eg, REDS4~\cite{nah2019ntire}) better than unidirectional propagation. Table~\ref{tab:quan} shows that both BasicVSR-uni and our CTUN can not obtain can not obtain satisfactory results on REDS4~\cite{nah2019ntire}.
However, compared with BasicVSR~\cite{chan2021basicvsr}, our CTUN-bi achieves comparable results on REDS4~\cite{nah2019ntire} and better performance on other benchmarks with fewer parameters and more efficiency.

\begin{figure*}[t]
	\footnotesize
	\begin{center}
		\begin{tabular}{c c c c c c c c}
			\multicolumn{3}{c}{\multirow{5}*[55pt]{\hspace{-2mm} \includegraphics[width=0.49\linewidth, height=0.268\linewidth]{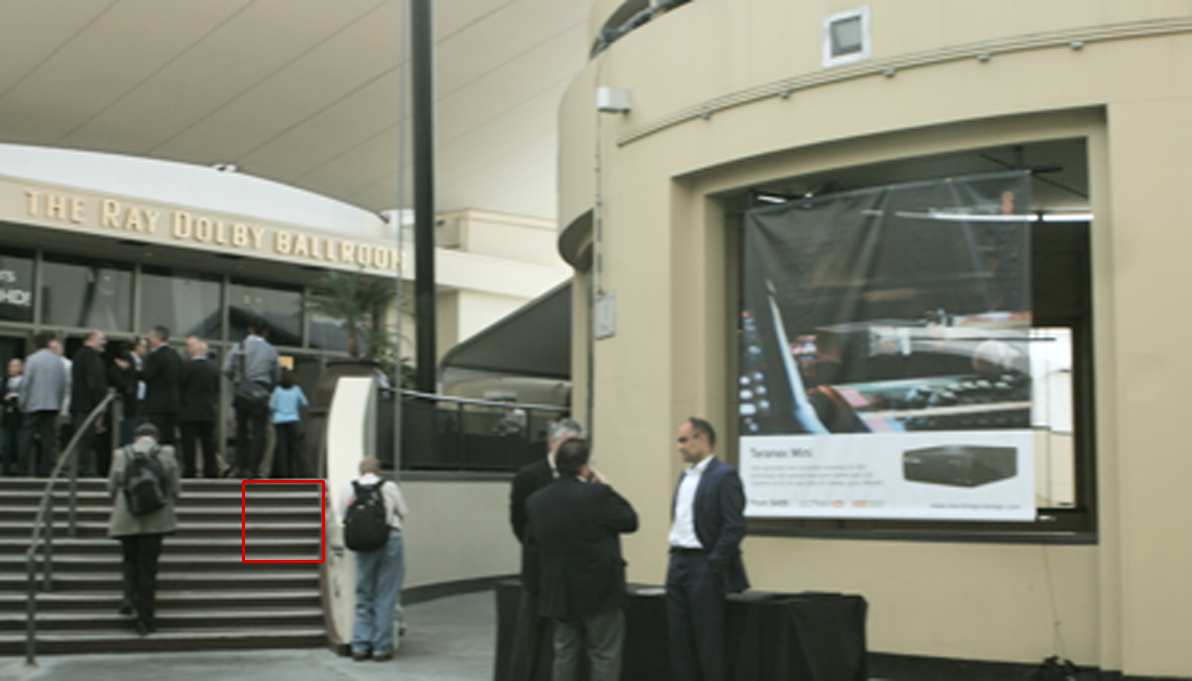}}}
            & \hspace{-4.0mm} \includegraphics[width=0.12\linewidth]{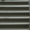} 
            & \hspace{-5.0mm} \includegraphics[width=0.12\linewidth]{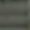}  
            & \hspace{-5.0mm}  \includegraphics[width=0.12\linewidth]{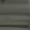}  
            & \hspace{-3.5mm} \includegraphics[width=0.12\linewidth]{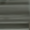}  \\
			\multicolumn{3}{c}{~}                                   &\hspace{-4.0mm} (a)  GT patch               & \hspace{-5.0mm} (b) Bicubic                    & \hspace{-5.0mm} (c) PFNL~\cite{yi2019progressive}  & \hspace{-3.5mm} (d) EDVR~\cite{wang2019edvr} \\		
			\multicolumn{3}{c}{~} 
            & \hspace{-4.0mm} \includegraphics[width=0.12\linewidth]{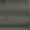} 
            & \hspace{-5.0mm} \includegraphics[width=0.12\linewidth]{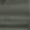} 
            & \hspace{-5.0mm} \includegraphics[width=0.12\linewidth]{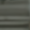} 
            & \hspace{-3.5mm} \includegraphics[width=0.12\linewidth]{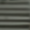}  \\
			\multicolumn{3}{c}{\hspace{-4.0mm} Sequence \textit{0216}, Clip \textit{024}} &\hspace{-4.0mm} (e) SSL~\cite{Xia2022ssl} & \hspace{-5.0mm} (f) BasicVSR-uni~\cite{chan2021basicvsr}  & \hspace{-5.0mm} (g) BasicVSR~\cite{chan2021basicvsr}          & \hspace{-3.5mm}(h) CTUN (Ours) \\			
		\end{tabular}
	\end{center}
	\vspace{-6mm}
	\caption{Visual comparisons on the Vimeo-T (BI) dataset. Only our method is able to reconstruct the stairs.}
	\label{fig:visual2}
	\vspace{-4mm}
\end{figure*}

\noindent \textbf{Qualitative comparisons.} Figures~\ref{fig:visual1} and~\ref{fig:visual2} present some visual quality comparisons, indicating the effectiveness of our method compared to other methods. 
As sliding windows-based methods~\cite{yi2019progressive,wang2019edvr} do not effectively explore long-range temporal information, the super-resolved results by these methods~\cite{yi2019progressive,wang2019edvr} contain significant blur effect (see Figure~\ref{fig:visual1}(c) and (d)).
This phenomenon is also observed in the unidirectional propagation method `BasicVSR-uni' (see Figure~\ref{fig:visual1}(f)). 
Although the bidirectional propagation methods~\cite{Xia2022ssl,chan2021basicvsr} based on optical flow alignment partially mitigate blurring, they struggle to accurately restore the structure information of the panes. (see Figure~\ref{fig:visual1}(e) and (g)).
Compared to other methods, our proposed ICAM is able to extract the most relevant spatio-temporal correspondences from adjacent frames, resulting in clearer structure information (see Figure~\ref{fig:visual1}(h)).
Also, Figure~\ref{fig:visual2} demonstrates that only our proposed method is capable of effectively restoring the details information of the stairs, while other methods exhibit various blurring and artifacts.
More visual comparisons can be found in the supplementary material.

\begin{figure}[t]
\footnotesize
\centering
    \begin{tabular}{ccc}
    \hspace{-2mm} \includegraphics[width=0.15\textwidth]{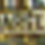}
    & \hspace{-4mm} \includegraphics[width=0.15\textwidth]{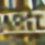}
    & \hspace{-3.5mm} \includegraphics[width=0.15\textwidth]{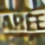} \\
     \hspace{-2mm} (a) Bicubic & \hspace{-4mm} (b) w/o ICAM \& HU   & \hspace{-3.5mm}(c) w/ ICAM \& w/o HU  \\
     \hspace{-2mm} \includegraphics[width=0.15\textwidth]{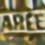}
    & \hspace{-4mm} \includegraphics[width=0.15\textwidth]{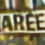}
    &\hspace{-3.5mm} \includegraphics[width=0.15\textwidth]{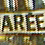} \\
     \hspace{-2mm} (d) w/o ICAM \& w/ HU    & \hspace{-4mm} (e) CTUN    & \hspace{-3.5mm} (f) GT 
    \end{tabular}
\vspace{-3mm}
    \caption{Effectiveness of our proposed module for VSR.}
    \label{fig:ablation1}
\vspace{-7mm}
\end{figure}

\noindent \textbf{Efficiency comparisons.} To fully examine the effiency of our proposed method, we evaluate our method against other methods in terms of the network parameters, FLOPs and the running time (see Table~\ref{tab:quan}).
The FLOPs are calculated on three $180\times320$ input frames, and the running time is averaged on 100 input frames with the $180\times320$ resolution.
Our method achieves the best result in terms of the running time, which is $3\times$ faster than BasicVSR~\cite{chan2021basicvsr}.
Moreover, as we utilize the proposed lightweight ICAM rather than other alignment module with high parameters (\eg, optical flow, deformable convolution) for alignment, our method achieves the second-best in terms of network parameters.
However, the method with the lowest parameters, TOFlow~\cite{xue2019video}, exhibits a running time nearly $80 \times$ longer than ours.
Although SSL~\cite{Xia2022ssl} exhibits efficiency metrics similar to our method, due to the limitations of structured pruning, our method outperforms it in terms of PSNR and SSIM on all benchmarks.

\vspace{-2mm}
\section{Analysis and Discussion}
\vspace{-2mm}
To demonstrate the effectiveness and efficiency of our proposed method, we conduct extensive ablation studies and provide deeper analysis in this session.
For all ablation studies in this section, we train the proposed method and all the baselines on the Vimeo-90K~\cite{xue2019video} (BI) dataset and evaluate on the Vid4~\cite{liu2014bayesian} (BI) dataset for fair comparison.

\noindent \textbf{Effectiveness of the ICAM.} The proposed ICAM is used to explore spatio-temporal correspondences between the current frame and adjacent frames for alignment.
To examine whether it facilitates efficient VSR, we first compare the proposed CTUN with a baseline that removes the ICAM and train this baseline under the same settings as ours for fairness.
Table~\ref{tab:ablation} shows that the proposed CTUN achieves a PSNR gain of 0.30dB with only 2ms increase in the running time compared with the baseline without using the ICAM (`w/o ICAM \& w/ HU').
Moreover, we train the baseline that removes all the proposed components (`w/o ICAM \& HU') and the baseline only with the ICAM (`w/ ICAM \& w/o HU').
The comparisons in Table~\ref{tab:ablation} illustrate the significance of using the proposed ICAM.
These quantitative comparisons indicate that the proposed ICAM can bring significant improvements without compromising the efficiency of the model.

As the proposed ICAM is used to align the current frame, one may wonder whether other alignment modules generate better results or not.
To answer this question, we use the alignment modules in~\cite{chan2021basicvsr} and \cite{chan2022basicvsrplusplus}, \ie, SpyNet and Deformable Alignment, to replace ICAM and conduct ablation studies.
Table~\ref{tab:ICAM} shows that using the alignment module based optical flow, \ie, SpyNet, does not achieve favorable results but leads to nearly double the network parameter.
Although using the deformable alignment~\cite{chan2022basicvsrplusplus} can obtain similar results with our ICAM, it significantly increases the model complexity.

\begin{figure}[t]
\footnotesize
\centering
    \begin{tabular}{cccc}
    \hspace{-2mm} \includegraphics[width=0.11\textwidth]{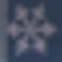} 
    & \hspace{-3.7mm} \includegraphics[width=0.11\textwidth]{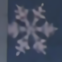} 
    & \hspace{-5.1mm} \includegraphics[width=0.11\textwidth]{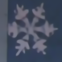} 
    & \hspace{-5.2mm} \includegraphics[width=0.11\textwidth]{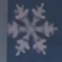} \\
    \hspace{-2mm}(a) Bicubic &\hspace{-3.7mm} (b) SpyNet~\cite{ranjan2017optical} &\hspace{-5.1mm} (c) Defor.Align.~\cite{chan2022basicvsrplusplus} &\hspace{-5.2mm} (d) Ours\\
    \end{tabular}
\vspace{-3mm}
    \caption{Effectiveness of the proposed ICAM for VSR.}
    \label{fig:ablation2}
\vspace{-2.41mm}
\end{figure}

For qualitative comparisons, Figure~\ref{fig:ablation1} shows that the method without using ICAM and HU (`w/o ICAM \& HU') contains significant blur effect on the text, while the method using the ICAM (`w/ ICAM \& w/o HU') can restore more texture information.
And compared with our method without using the ICAM (`w/o ICAM \& w/ HU'), our CTUN is able to generate much clearer texture and edge information.
Furthermore, Figure~\ref{fig:ablation2} illustrates that our proposed ICAM can restore better structure details of the snowflake compared to using SpyNet~\cite{ranjan2017optical} and deformable alignment~\cite{chan2022basicvsrplusplus}.

\begin{table}[t]
    \caption{Quantitative evaluations of the proposed ICAM and HU on the Vid4 dataset.}
    \vspace{-6mm}
    \label{tab:ablation}
    \small
    \begin{center}
        \tabcolsep=0.3cm
        \scalebox{0.62}{
            \begin{tabular}{lcccc}
                \toprule
                Methods            & w/o ICAM \& HU   & w/ ICAM \& w/o HU   & w/o ICAM \& w/ HU   & CTUN      \\ \hline 
                Parameters (M)    & 1.30            & 1.70                  & 1.89                  & 2.25                \\
                Runtimes (ms)    & 13                  & 16                  & 19                  & 21                \\ 
                PSNR (dB)         & 26.72               & 27.13                 & 27.18                 & \textbf{27.48}      \\
                \bottomrule
            \end{tabular}}
    \end{center}
    \vspace{-9mm}
\end{table}

\noindent \textbf{Effectiveness of the HU.} The proposed HU module is used to introduce future information in unidirectional propagation, aiming to compensate for the performance gap compared to bidirectional propagation.
To demonstrate its effectiveness on VSR, we compare the proposed method in two ways.
First, Table~\ref{tab:ablation} shows that our CTUN and the method using the proposed HU (`w/o ICAM \& w/ HU') achieve improvements of at least 0.35dB in terms of PSNR compared to methods without using the HU (`w/o ICAM \& HU' and `w/ ICAM \& w/o HU'), respectively.

In addition, as the proposed HU module mainly contains a feature encoder~\eqref{eq:CBAM} and an updating GRU~\eqref{eq:update}, we conduct ablation studies \wrt these components to demonstrate their effectiveness on VSR.
The feature encoder~\eqref{eq:CBAM} is used to explore spatial information from concatenated features.
The `w/o~\eqref{eq:CBAM}' in Table~\ref{tab:HU} represents that we replace the feature encoder with a $3\times3$ convolution to extract spatial information, and the results in Table~\ref{tab:HU} show that our method using \eqref{eq:CBAM} obtains better PSNR value. 

The updating GRU~\eqref{eq:update} aims to update hidden features $h_t$ by achieving temporal interaction between $m_t$ and $h_t$.
We first remove \eqref{eq:update} and directly use $m_t$ obtained by \eqref{eq:CBAM} to update $h_t$ (see \eqref{eq:newforward}), denote as `w/o \eqref{eq:update}'.
And Table~\ref{tab:HU} represents that it brings a decrease of 0.13 dB in PSNR, indicating that the temporal interaction between $m_t$ and $h_t$ is crucial for VSR.
Furthermore, we remove the three components $\{m^p,m^c,m^u\}$ in \eqref{eq:update}, and \eqref{eq:update} changes as follows:
\begin{equation}
    \begin{split}
        &z = \sigma(\text{Conv}_{3 \times 3}(m_t)), \\
        &w = \sigma(\text{Conv}_{3 \times 3}(m_t)), \\
        &q = \text{tanh}(\text{Conv}_{3 \times 3}(m_t)),\\ 
        &m_t = z \otimes h_t \otimes (1-w) + q \otimes w.
    \end{split}
    \label{eq:update_2}
\end{equation}

Table~\ref{tab:HU} shows that our method achieves an improvement of 0.11dB in terms of PSNR compared with `w/o $\{m^p,m^c,m^u\}$', demonstrating that using the cascaded temporal information in updating GRU is necessary.

\begin{table}[t]
    \caption{Effect of the proposed ICAM on the Vid4 dataset.}
    \centering
    \vspace{-2.5mm}
    \resizebox{0.47\textwidth}{!}{
    \begin{tabular}{lccc}
        \toprule
         Methods & SpyNet~\cite{ranjan2017optical} & Deformable Alignment~\cite{chan2022basicvsrplusplus} & Ours \\ \hline
         Parameters (M)  & 3.37            & 5.39                  & 2.25               \\
         PSNR (dB)   & 27.21  & 27.44  & \textbf{27.48}    \\ 
         \bottomrule
    \end{tabular}
    }
    \label{tab:ICAM}
    \vspace{-2mm}
\end{table}

\begin{figure}[t]
    \centering
    \includegraphics[width=0.45\textwidth]{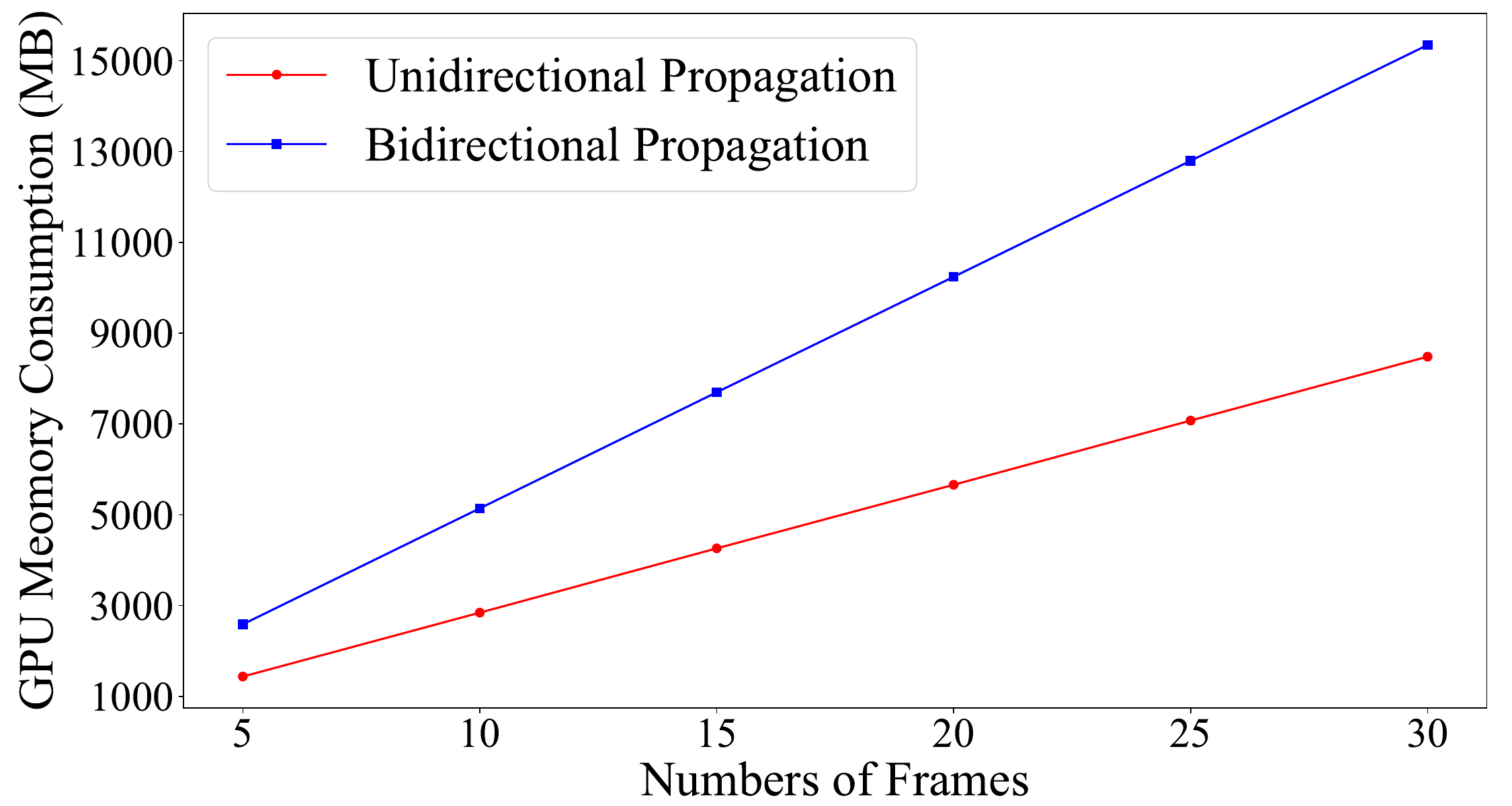}
    \vspace{-2mm}
    \caption{GPU memory consumption comparisons during the inference phase. The spatial size of input frames is $64 \times 64$.}
    \label{fig:memory}
    \vspace{-6mm}
\end{figure}

\begin{figure}[t]
\footnotesize
\centering
    \begin{tabular}{cccc}
    \hspace{-2mm} \includegraphics[width=0.11\textwidth]{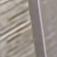} 
    & \hspace{-3.7mm} \includegraphics[width=0.11\textwidth]{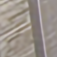} 
    & \hspace{-7.8mm} \includegraphics[width=0.11\textwidth]{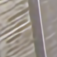} 
    & \hspace{-7.5mm} \includegraphics[width=0.11\textwidth]{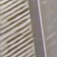} \\
    \hspace{-2mm}(a) w/o \eqref{eq:CBAM} &\hspace{-3.7mm} (b) w/o \eqref{eq:update} &\hspace{-7.8mm} (c) w/o $\{m^p,m^c,m^u\}$ &\hspace{-7.5mm} (d) Ours\\
    \end{tabular}
\vspace{-3mm}
    \caption{Effectiveness of the proposed HU for VSR.}
    \label{fig:ablation3}
\vspace{-3.5mm}
\end{figure}

For qualitative comparisons, Figure~\ref{fig:ablation1} presents that the models without using the HU do not produce satisfactory results, while the proposed method using the HU generates much clearer structural and edge information.
Figure~\ref{fig:ablation3} further demonstrates that both the spatial information by \eqref{eq:CBAM} and the temporal interaction by \eqref{eq:update} are useful for clear video reconstruction.

\noindent \textbf{GPU memory consumption.} To further examine the efficiency of our proposed method, we evaluate the GPU memory consumption of bidirectional and unidirectional propagation networks for video sequences of different lengths during the inference, with ~\href{https://pytorch.org/docs/stable/generated/torch.cuda.max\_memory\_allocated.html}{torch.cuda.max\_memory\_allocated()} function. Figure~\ref{fig:memory} presents that as the number of input frames increases, the GPU memory required by bidirectional propagation grows faster than unidirectional propagation, indicating that the proposed unidirectional propagation updated network is more friendly to resource-limited devices.

\begin{figure}[t]
    \centering
    \includegraphics[width=0.45\textwidth]{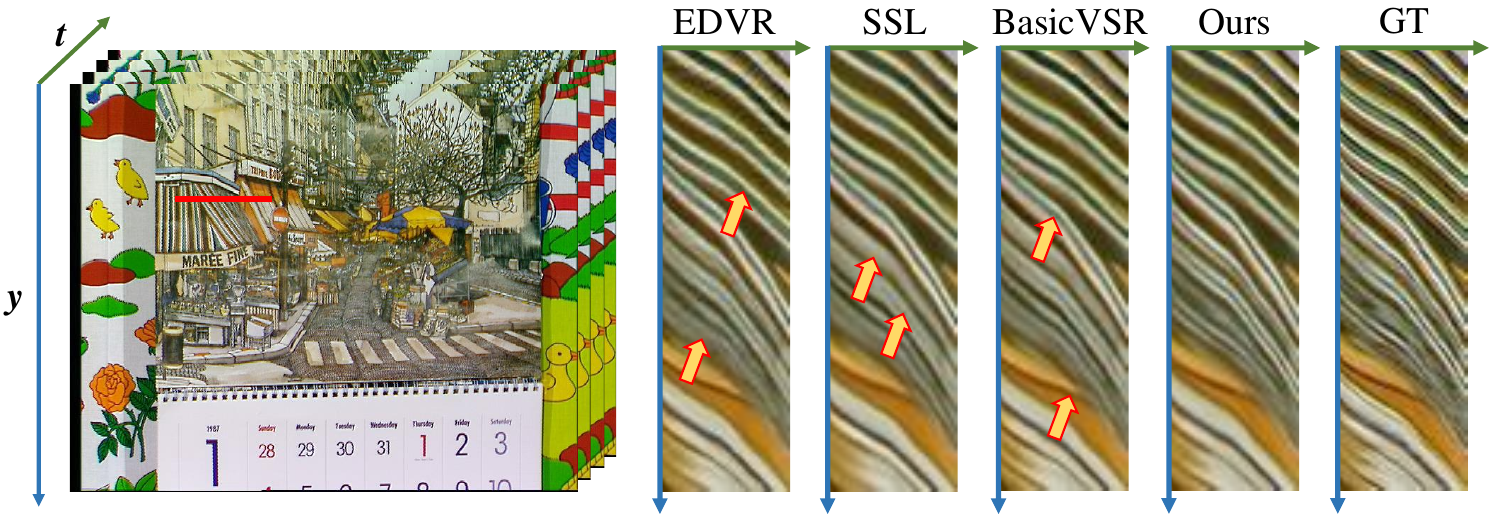}
    \caption{Visual comparisons of the temporal consistency for reconstructed videos. We visualize the pixels of the \rf{red} line.}
    \label{fig:Temporal_consistency}
    \vspace{-6mm}
\end{figure}

\noindent \textbf{Temporal consistency property.} Following~\cite{chan2022basicvsrplusplus}, we further evaluate the temporal consistency property of the reconstructed videos.
Figure~\ref{fig:Temporal_consistency} presents the temporal profile to qualitatively compare temporal consistency, where our method exhibits smoother temporal transition.


\begin{table}[t]
    \caption{Effect of the proposed HU on the Vid4 dataset.}
    \centering
    \vspace{-2.5mm}
    \resizebox{0.47\textwidth}{!}{
    \begin{tabular}{lcccc}
        \toprule
         Methods & w/o~\eqref{eq:CBAM} & w/o~\eqref{eq:update} & w/o $\{m^p,m^c,m^u\}$ & Ours \\ \hline
         Parameters (M)  & 1.93            & 2.07          & 2.23      & 2.25                \\
         PSNR (dB)  & 27.40  & 27.35  & 27.37  & \textbf{27.48} \\ 
         \bottomrule
    \end{tabular}
    }
    \label{tab:HU}
    \vspace{-2mm}
\end{table}

\vspace{-2mm}
\section{Conclusion}
\vspace{-2mm}
In this paper, we have presented an effective lightweight cascaded temporal propagation network with an updating mechanism, called CTUN, for efficient video super-resolution.
We first develop an implicit cascaded alignment module to effectively explore the spatio-temporal correspondences and efficiently align the current frames with its adjacent frames in a cascaded manner.
To explore long-range temporal information efficiently, we develop a unidirectional propagation network that incorporates a simple yet effective hidden updating GRU module, enabling the introduction of future information in unidirectional propagation.
We formulate each component into an end-to-end trainable deep model and show that our model is more compact and efficient for video super-resolution.
We conduct extensive experiments and analysis to demonstrate the effectiveness of our method.
Both quantitative and qualitative results show that the proposed method performs favorably against state-of-the-art methods in terms of accuracy and model complexity.

{
    \small
    \bibliographystyle{ieeenat_fullname}
    \bibliography{main}
}

\end{document}